# *Roget's Thesaurus* and Semantic Similarity

**Mario JARMASZ and Stan SZPAKOWICZ**
School of Information Technology
and Engineering
University of Ottawa
Ottawa, Ontario, Canada, K1N 6N5
{mjarmasz,szpak}@site.uottawa.ca

## Abstract

We have implemented a system that measures semantic similarity using a computerized 1987 *Roget's Thesaurus*, and evaluated it by performing a few typical tests. We compare the results of these tests with those produced by *WordNet*-based similarity measures. One of the benchmarks is Miller and Charles' list of 30 noun pairs to which human judges had assigned similarity measures. We correlate these measures with those computed by several NLP systems. The 30 pairs can be traced back to Rubenstein and Goodenough's 65 pairs, which we have also studied. Our *Roget*'s-based system gets correlations of .878 for the smaller and .818 for the larger list of noun pairs; this is quite close to the .885 that Resnik obtained when he employed humans to replicate the Miller and Charles experiment. We further evaluate our measure by using *Roget's* and *WordNet* to answer 80 *TOEFL*, 50 *ESL* and 300 *Reader's Digest* questions: the correct synonym must be selected amongst a group of four words. Our system gets 78.75%, 82.00% and 74.33% of the questions respectively.

## 1 Introduction

People identify synonyms — strictly speaking, near-synonyms (Edmonds and Hirst, 2002) — such as *angel – cherub*, without being able to define synonymy properly. The term tends to be used loosely, even in the crucially synonymy-oriented *WordNet* with the *synset* as the basic semantic unit (Fellbaum, 1998, p. 23). Miller and Charles (1991) restate a formal, and linguistically quite inaccurate, definition of synonymy usually attributed to Leibniz: "two words are said to be synonyms if one can be used in a statement in place of the other without changing the meaning of the statement". With this strict definition there may be no perfect synonyms in natural language (Edmonds and Hirst, *ibid.*). For NLP systems it is often more useful to establish the degree of synonymy between two words, referred to as *semantic similarity*.

Miller and Charles' semantic similarity is a continuous variable that describes the degree of synonymy between two words (*ibid.*). They argue that native speakers can order pairs of words by semantic similarity, for example *ship – vessel*, *ship – watercraft*, *ship – riverboat*, *ship – sail*, *ship – house*, *ship – dog*, *ship – sun*. The concept can be usefully extended to quantify relations between non-synonymous but closely related words, for example *airplane – wing*.

Rubenstein and Goodenough (1965) investigated the validity of the assumption that "... pairs of words which have many contexts in common are semantically closely related". This led them to establish *synonymy judgments* for 65 pairs of nouns with the help of human experts. Miller and Charles (*ibid.*) selected 30 of those pairs, and studied semantic similarity as a function of the contexts in which words are used. Others have calculated similarity using semantic nets (Rada *et al.*, 1989), in particular *WordNet* (Resnik, 1995; Jiang and Conrath, 1997; Lin, 1998; Hirst and St-Onge, 1998; Leacock and Chodorow, 1998) and *Roget's Thesaurus* (McHale, 1998), or statistical methods (Landauer and Dumais, 1997; Turney, 2001; Bigham *et al.*, 2003)

We set out to test the intuition that *Roget's Thesaurus*, sometimes treated as a book of synonyms, allows us to measure semantic similarity effectively. We demonstrate some of *Roget's* qualities which make it a realistic alternative to *WordNet*, in particular for the task of measuring semantic similarity. We propose a measure of *semantic distance*, the inverse of semantic similarity (Budanitsky & Hirst 2001) based on *Roget's* taxonomy. We convert it into a semantic similarity measure, and empirically compare to human judgments and to those of NLP systems. We consider the tasks of assigning a similarity value to pairs of nouns and choosing the correct synonym of a problem word given the choice of four target words. We explain in detail the measures and the experiments, and draw a few conclusions.

## 2 *Roget's Thesaurus* Relations as a Measure of Semantic Distance

Resnik (1995) claims that a natural way of calculating semantic similarity in a taxonomy is to measure the distance between the nodes that correspond to the items we compare: the shorter the path, the more similar the items. Given multiple paths, we take the length of the shortest one. Resnik states a widely acknowledged problem with edge counting. It relies on the notion that links in the taxonomy represent uniform distances, and it is therefore not the best semantic distance measure for *WordNet*. We want to investigate this claim for *Roget's*, as its hierarchy is very regular.

*Roget's Thesaurus* has many advantages. It is based on a well-constructed concept classification, and its entries were written by professional lexicographers. It contains around 250,000 words compared to *WordNet's* almost 200,000. *Roget's* does not have some of *WordNet's* shortcomings, such as the lack of links between parts of speech and the absence of topical groupings. The clusters of closely related words are obviously not the same in both resources. *WordNet* relies on a set of about 15 semantic relations. Search in this lexical database requires a word and a semantic relation; for every word some (but never all) of 15 relations can be used in search. It is impossible to express a relationship that involves more than one of the 15 relations: it cannot be stored in *WordNet*. The *Thesaurus* can link the noun *bank*, the business that provides financial services, and the verb *invest*, to give money to a bank to get a profit, as used in the following sentences, by placing them in a common head **784** *Lending*.

1. Mary went to the *bank* yesterday.
2. She *invested* $5,000.00 in mutual funds.

This type of connection cannot be described using *WordNet's* semantic relations. While an English speaker can identify a relation not provided by *WordNet*, for example that one invests money in a bank, this is not sufficient for use in computer systems.

We used a computerized version of the 1987 edition of Penguin's *Roget's Thesaurus of English Words and Phrases* (Jarmasz and Szpakowicz, 2001) to calculate the semantic distance. *Roget's* structure allows an easy implementation of edge counting. Given two words, we look up in the index their references that point into the *Thesaurus*. Next, we calculate all paths between references using *Roget's* taxonomy. Using another version of *Roget's*, McHale (1998) showed that edge counting is a good semantic distance measure.

Eight Classes head this taxonomy. The first three, *Abstract Relations, Space* and *Matter*, cover the external world. The remaining ones, *Formation of ideas, Communication of ideas, Individual volition, Social volition, Emotion, Religion and Morality* deal with the internal world of human beings. A path in *Roget's* ontology always begins with one of the Classes. It branches to one of the 39 Sections, then to one of the 79 Sub-Sections, then to one of the 596 Head Groups and finally to one of the 990 Heads. Each Head is divided into paragraphs grouped by parts of speech: nouns, adjectives, verbs and adverbs. Finally a paragraph is divided into semicolon groups of semantically closely related words. Jarmasz and Szpakowicz (*ibid.*) give a detailed account of *Roget's* structure.

The distance equals the number of edges in the shortest path. Path lengths are as follows.

- Length 0: the same semicolon group.
  *journey's end – terminus*
- Length 2: the same paragraph.
  *devotion – abnormal affection*
- Length 4: the same part of speech.
  *popular misconception – glaring error*
- Length 6: the same head.
  *individual – lonely*
- Length 8: the same head group.
  *finance – apply for a loan*
- Length 10: the same sub-section.
  *life expectancy – herbalize*
- Length 12: the same section.
  *Creirwy (love) – inspired*
- Length 14: the same class.
  *translucid – blind eye*
- Length 16: in the *Thesaurus*.
  *nag – like greased lightning*

As an example, the *Roget's* distance between *feline* and *lynx* is 2. The word **feline** has these references:

  1) *animal 365 ADJ.*
  2) *cat 365 N.*
  3) *cunning 698 ADJ.*

The word **lynx** has these references:

  1) *cat 365 N.*
  2) *eye 438 N.*

The shortest and the longest path are:

- *feline → cat ← lynx*
- *feline → cunning → ADJ. → 698. Cunning → [698, 699] → Complex → Section three : Voluntary action → Class six : Volition: individual volition → T ← Class three : Matter ← Section three : Organic matter ← Sensation ← [438, 439, 440] ← 438. Vision ← N. ← eye ← lynx*

| Noun Pair | Miller Charles | Penguin Roget | WordNet Edges | Hirst St.Onge | Jiang Conrath | Leacock Chodorow | Lin | Resnik |
|---|---|---|---|---|---|---|---|---|
| *car – automobile* | 3.920 | 16.000 | 30.000 | 16.000 | 1.000 | 3.466 | 1.000 | 6.340 |
| *gem – jewel* | 3.840 | 16.000 | 30.000 | 16.000 | 1.000 | 3.466 | 1.000 | 12.886 |
| *journey – voyage* | 3.840 | 16.000 | 29.000 | 4.000 | 0.169 | 2.773 | 0.699 | 6.057 |
| *boy – lad* | 3.760 | 16.000 | 29.000 | 5.000 | 0.231 | 2.773 | 0.824 | 7.769 |
| *coast – shore* | 3.700 | 16.000 | 29.000 | 4.000 | 0.647 | 2.773 | 0.971 | 8.974 |
| *asylum – madhouse* | 3.610 | 16.000 | 29.000 | 4.000 | 0.662 | 2.773 | 0.978 | 11.277 |
| *magician – wizard* | 3.500 | 14.000 | 30.000 | 16.000 | 1.000 | 3.466 | 1.000 | 9.708 |
| *midday – noon* | 3.420 | 16.000 | 30.000 | 16.000 | 1.000 | 3.466 | 1.000 | 10.584 |
| *furnace – stove* | 3.110 | 14.000 | 23.000 | 5.000 | 0.060 | 1.386 | 0.238 | 2.426 |
| *food – fruit* | 3.080 | 12.000 | 23.000 | 0.000 | 0.088 | 1.386 | 0.119 | 0.699 |
| *bird – cock* | 3.050 | 12.000 | 29.000 | 6.000 | 0.159 | 2.773 | 0.693 | 5.980 |
| *bird – crane* | 2.970 | 14.000 | 27.000 | 5.000 | 0.139 | 2.079 | 0.658 | 5.980 |
| *tool – implement* | 2.950 | 16.000 | 29.000 | 4.000 | 0.546 | 2.773 | 0.935 | 5.998 |
| *brother – monk* | 2.820 | 14.000 | 29.000 | 4.000 | 0.294 | 2.773 | 0.897 | 10.489 |
| *lad – brother* | 1.660 | 14.000 | 26.000 | 3.000 | 0.071 | 1.856 | 0.273 | 2.455 |
| *crane – implement* | 1.680 | 0.000 | 26.000 | 3.000 | 0.086 | 1.856 | 0.394 | 3.443 |
| *journey – car* | 1.160 | 12.000 | 17.000 | 0.000 | 0.075 | 0.827 | 0.000 | 0.000 |
| *monk – oracle* | 1.100 | 12.000 | 23.000 | 0.000 | 0.058 | 1.386 | 0.233 | 2.455 |
| *cemetery – woodland* | 0.950 | 6.000 | 21.000 | 0.000 | 0.049 | 1.163 | 0.067 | 0.699 |
| *food – rooster* | 0.890 | 6.000 | 17.000 | 0.000 | 0.063 | 0.827 | 0.086 | 0.699 |
| *coast – hill* | 0.870 | 4.000 | 26.000 | 2.000 | 0.148 | 1.856 | 0.689 | 6.378 |
| *forest – graveyard* | 0.840 | 6.000 | 21.000 | 0.000 | 0.050 | 1.163 | 0.067 | 0.699 |
| *shore – woodland* | 0.630 | 2.000 | 25.000 | 2.000 | 0.056 | 1.674 | 0.124 | 1.183 |
| *monk – slave* | 0.550 | 6.000 | 26.000 | 3.000 | 0.063 | 1.856 | 0.247 | 2.455 |
| *coast – forest* | 0.420 | 6.000 | 24.000 | 0.000 | 0.055 | 1.520 | 0.121 | 1.183 |
| *lad – wizard* | 0.420 | 4.000 | 26.000 | 3.000 | 0.068 | 1.856 | 0.265 | 2.455 |
| *chord – smile* | 0.130 | 0.000 | 20.000 | 0.000 | 0.066 | 1.068 | 0.289 | 2.888 |
| *glass – magician* | 0.110 | 2.000 | 23.000 | 0.000 | 0.056 | 1.386 | 0.123 | 1.183 |
| *rooster – voyage* | 0.080 | 2.000 | 11.000 | 0.000 | 0.044 | 0.470 | 0.000 | 0.000 |
| *noon – string* | 0.080 | 6.000 | 19.000 | 0.000 | 0.052 | 0.981 | 0.000 | 0.000 |
| **Correlation** | 1.000 | 0.878 | 0.732 | 0.689 | 0.695 | 0.821 | 0.823 | 0.775 |

**Table 1:** Comparison of semantic similarity measures using the Miller and Charles data

| | Rubenstein Goodenough | Penguin Roget | WordNet Edges | Hirst St.Onge | Jiang Conrath | Leacock Chodorow | Lin | Resnik |
|---|---|---|---|---|---|---|---|---|
| **Correlation** | 1.000 | 0.818 | 0.787 | 0.732 | 0.731 | 0.852 | 0.834 | 0.800 |

**Table 2:** Comparison of semantic similarity measures using the Rubenstein and Goodenough data

We convert the distance measure to similarity by subtracting the path length from the maximum possible path length (Resnik, 1995):

$$\text{sim}(w_1, w_2) = 16 - [\min \text{distance}(r_1, r_2)] \quad (1)$$

where $r_1$ and $r_2$ are the sets of references for the words or phrases $w_1$ and $w_2$.

## 3 Evaluation Based on Human Judgment

### 3.1 The Data

Rubenstein and Goodenough (1965) established *synonymy judgments* for 65 pairs of nouns. They invited 51 judges who assigned to every pair a score between 4.0 and 0.0 indicating semantic similarity. They chose words from non-technical every day English. They felt that, since the phenomenon under investigation was a general property of language, it was not necessary to study technical vocabulary. Miller and Charles (1991) repeated the experiment restricting themselves to 30 pairs of nouns selected from Rubentein and Goodenough's list, divided equally amongst words with high, intermediate and low similarity.

We repeated both experiments using the *Roget's Thesaurus* system. We decided to compare our results to six other similarity measures that rely on *WordNet*. Pedersen's Semantic Distance software package (2002) was used with *WordNet 1.7.1* to obtain the results. The first *WordNet* measure used is edge counting. It serves as a baseline, as it is the simplest and most intuitive measure. The nextThe next measure, from Hirst and St-Onge (1998), relies on the path length as well as the number of changes of direction in the path; these changes are defined in function of *WordNet* semantic relations. Jiang and Conrath (1997) propose a combined approach based on edge counting enhanced by the node-based approach of the information content calculation proposed by Resnik (1995). Leacock and Chodorow (1998) count the path length in nodes rather than links, and adjust it to take into account the maximum depth of the taxonomy. Lin (1998) calculates semantic similarity using a formula derived from information theory. Resnik (1995) calculates the information content of the concepts that subsume them in the taxonomy. We calculate the Pearson product-moment correlation coefficient between the human judgments and the values achieved by the systems. The correlation is significant to at the 0.01 level. These similarity measures appear in Tables 1 and 2.

### 3.2 The Results

We begin by analyzing the results obtained by *Roget's*. The Miller and Charles data in Table 1 show that pairs of words with a semantic similarity value of 16 have high similarity, those with a score of 12 to 14 have intermediate similarity, and those with a score below 10 are of low similarity. This is intuitively correct, as words or phrases that are in the same semicolon group will have a similarity score of 16, those that are in the same paragraph, part-of-speech or head will have a score of 10 to 14, and words that cannot be found in the same head, therefore do not belong to the same concept, will have a score between 0 and 8. *Roget's* results correlate very well with human judgment for the Miller and Charles list ($r=.878$), almost attaining the upper bound ($r=.885$) set by human judges (Resnik, 1995) despite the outlier *crane – implement*, two words that have nothing in common in the *Thesaurus*.

The correlation between human judges and *Roget's* for the Rubenstein and Goodenough data is also very good ($r=.818$) as shown in Table 2. Although we do not present the 65 pairs of words in the list, the outliers merit discussion. Five pairs of low similarity words are deemed to be of intermediate similarity by *Roget's*, all with the semantic distance value of 12. These pairs of words are therefore all found under the same Head and belong to noun groups. The associations made by the *Thesaurus* are correct but not the most intuitive: *glass - jewel* is assigned a value of 1.78 by the human judges but can be found under the Head *844 Ornamentation*, *car – journey* is assigned 1.55 and is found under the Head *267 Land travel*, *monk – oracle* 0.91 found under Head *986 Clergy*, *boy – rooster* 0.44 under Head *372 Male*, and *fruit – furnace* 0.05 under Head *301 Food: eating and drinking*.

The results might suggest that a *Roget's*-based measure will not scale up to larger sets of nouns. We repeated our experiment with a list of 353 word pairs assembled by Gabrilovich (2002). The correlation with human judges is a rather low .539, but is still better than the best *WordNet*-based score of .375, obtained using Resnik's function, and comparable to Finkelstein *et al.*'s (2002) combined metric which obtains a score of $r=.550$. We cannot simply attribute the low scores to the measures not scaling up to larger data sets. The Finkelstein *et al.* list contains pairs that are associated but not similar in the semantic sense, for example: *liquid – water*. The list also contains many culturally biased pairs, for example: *Arafat – terror*. The authors of the list say that it represents various degrees of similarity.

|  | *Hirst St-Onge* | *Jiang Conrath* | *Leacock Chodorow* | *Lin* | *Resnik* |
|---|---|---|---|---|---|
| *Original results* | N. / A. | 0.828 | 0.740 | 0.834 | 0.791 |
| *Budanistky Hirst* | 0.744 | 0.850 | 0.816 | 0.829 | 0.774 |
| *Pedersen Distance* | 0.689 | 0.696 | 0.832 | 0.846 | 0.787 |

**Table 3:** Comparison of correlation values for the different measures using the Miller and Charles data

They employed 16 subjects to rate the semantic similarity on a scale from 0 to 10, 0 representing totally unrelated words and 10 very much related or identical words (Finkelstein *et al.*, *ibid.*). They do not explain the methodology used for preparing this list. Human subjects find it more difficult to use a scale from 0 to 10 rather than a more typical one such as 0 to 4. These issues cast a doubt on the validity of this list, and we therefore do not consider it as a suitable benchmark for performing experiments on semantic similarity.

Resnik argues that edge counting using WordNet 1.4 is not a good measure of semantic similarity as it relies on the notion that links in the taxonomy represent uniform distances. Tables 1 and 2 show that this measure performs well for WordNet 1.7.1 . This could be explained by the substantial improvement in the newest version of *WordNet*, including more uniform distances between words.

Table 3 shows that it is difficult to replicate accurately experiments that use WordNet-based measures. Budanitsky and Hirst (2001) repeated the Miller and Charles experiment using the WordNet similarity measures of Hirst and St-Onge (1998), Jiang and Conrath (1997), Leacock and Chodorow (1998), Lin (1998) and Resnik (1995). They claim that the discrepancies in the results can be explained by minor differences in implementation, different versions of WordNet, and differences in the corpora used to obtain the frequency data used by the similarity measures. There are also discrepancies with the results obtained by Pedersen's software (2002). We concur with Budanitsky and Hirst, pointing out that the Resnik, Leacock and Chodorow as well as the Lin experiments were performed not using the entire Miller and Charles set, but a 28 noun-pair subset, as the word woodland was not in WordNet when they performed their experiments.

## 4 Evaluation Based on Synonymy Problems

### 4.1 The Data

Another method of evaluating semantic similarity metrics is to see how well a computer system can score on a standardized synonym test. Such tests have questions where the correct synonym is one of four possible choices. This type of questions can be found in the Test of English as a Foreign Language [*TOEFL*] (Landauer and Dumais, 1997) and English as a Second Language tests [*ESL*] (Turney, 2001), as well as the Reader's Digest Word Power Game [*RDWP*] (Lewis, 2000-2001). Although this evaluation method is not widespread in Computational Linguistics, it has been used in Psychology (Landauer and Dumais, *ibid.*) and Machine Learning (Turney, *ibid.*). In this experiment we use 80 *TOEFL*, 50 *ESL* and 300 *RDWP* questions.

A *RDWP* question is presented like this: "Check the word or phrase you believe is nearest in meaning. **ode** – A: heavy debt. B: poem. C: sweet smell. D: surprise." (Lewis, 2001, n. 938). Our system calculates the semantic distance between the problem word and each choice word or phrase. The choice word with the shortest semantic distance becomes the solution. Choosing the word or phrase that has the most paths with the shortest distance breaks ties. Phrases that cannot be found in the *Thesaurus* present a special problem. We calculate the distance between each word in the choice phrase and the problem word; the conjunction *and*, the preposition *to*, the verb *be* are ignored. The shortest distance between the individual words of the phrase and the problem word is considered as the semantic distance for the phrase. This technique, although simplistic, lets us deal with phrases like *rise and fall*, *to urge* and *be joyous* that may not be found in the *Thesaurus* as presented. The *Roget's* system is not restricted to nouns when finding the shortest path – nouns, adjectives, verbs and adverbs are all considered. Using the previous *RDWP* example, the system would output the following:

- *ode* N. to *heavy* N., length = 12, 42 path(s) of this length
- *ode* N. to *poem* N., length = 2, 2 path(s) of this length

|  | Penguin Roget | WordNet Edges | Hirst St.Onge | Jiang Conrath | Leacock Chodorow | Lin | Resnik | PMI-IR | LSA |
|---|---|---|---|---|---|---|---|---|---|
| *Correct* | 63 | 17 | 57 | 20 | 17 | 19 | 15 | 59 | 50 |
| *Questions with ties* | 0 | 1 | 18 | 0 | 1 | 1 | 3 | 0 | 6 |
| *Score* | 63 | 17.5 | 62.33 | 20 | 17.5 | 19.25 | 16.25 | 59 | 51.5 |
| *Percent* | **78.75** | **21.88** | **77.91** | **25.00** | **21.88** | **24.06** | **20.31** | **73.75** | **64.38** |
| *Questions not found* | 4 | 53 | 2 | 53 | 53 | 53 | 53 | 0 | 0 |
| *Other words not found* | 22 | 24 | 2 | 24 | 24 | 24 | 24 | 0 | 0 |

**Table 4:** Comparison of the similarity measures for answering 80 TOEFL questions

|  | Penguin Roget | WordNet Edges | Hirst St.Onge | Jiang Conrath | Leacock Chodorow | Lin | Resnik | PMI-IR |
|---|---|---|---|---|---|---|---|---|
| *Correct* | 41 | 16 | 29 | 18 | 16 | 18 | 15 | 37 |
| *Questions with ties* | 0 | 4 | 5 | 0 | 4 | 0 | 3 | 0 |
| *Score* | 41 | 18 | 31 | 18 | 18 | 18 | 16.33 | 37 |
| *Percent* | **82.00** | **36.00** | **62.00** | **36.00** | **36.00** | **36.00** | **32.66** | **74.00** |
| *Questions not found* | 0 | 11 | 0 | 11 | 11 | 11 | 11 | 0 |
| *Other words not found* | 2 | 23 | 2 | 23 | 23 | 23 | 23 | 0 |

**Table 5:** Comparison of the similarity measures for answering 50 ESL questions

|  | Penguin Roget | WordNet Edges | Hirst St.Onge | Jiang Conrath | Leacock Chodorow | Lin | Resnik |
|---|---|---|---|---|---|---|---|
| *Correct* | 223 | 68 | 123 | 68 | 68 | 63 | 59 |
| *Questions with ties* | 0 | 3 | 44 | 1 | 3 | 9 | 14 |
| *Score* | 223 | 69.33 | 136.92 | 68.5 | 69.33 | 66.17 | 64 |
| *Percent* | **74.33** | **23.11** | **45.64** | **22.83** | **23.11** | **22.06** | **21.33** |
| *Questions not found* | 21 | 114 | 6 | 114 | 114 | 114 | 114 |
| *Other words not found* | 18 | 340 | 377 | 340 | 340 | 340 | 340 |

**Table 6:** Comparison of the similarity measures for answering 300 Reader's Digest questions

- *ode* N. to *sweet smell* N., length = 16, 6 path(s) of this length
- *ode* N. to *surprise* VB., length = 12, 18 path(s) of this length

→ Roget thinks that ode means poem: **CORRECT**

Note that the shortest distance between *ode* and *heavy debt* is that between *ode* and *heavy*.

We put the *WordNet* semantic similarity measures to the same task of answering the synonymy questions. The purpose of our experiment was not to improve the measures, but to use them as a comparison for the *Roget's* system. We choose as the answer the choice word that has the largest semantic similarity value with the problem word. When ties occur, a partial score is given; .5 if two words are tied for the highest similarity value, .33 if three, and .25 if four. The results appear in Tables 4-6. We did not tailor the *WordNet* measures to the task of answering these questions. All of them, except Hirst and St-Onge, rely on the IS-A hierarchy to calculate the path between words. The measures have been limited to finding similarities between nouns, as the *WordNet* hyponym tree only exists for nouns and verbs; there are hardly any links between parts of speech. We did not implement any special techniques to deal with phrases. It is therefore quite probable that the similarity measures can be improved for the task of answering synonymy questions.

We also compare our results to those achieved by state-of-the-art statistical techniques. Latent Semantic Analysis [*LSA*] is a general theory of acquired similarity and knowledge representation (Landauer and Dumais, 1997). It was used to answer the 80 *TOEFL* questions. The algorithm, called *PMI-IR* (Turney, 2001), uses Pointwise Mutual Information [*PMI*] and Information Retrieval [*IR*] to measure the similarity of pairs of words. It has been evaluated using the *TOEFL* and *ESL* questions. Bigham *et al.* (2003) combine four statistical methods, including *LSA* and *PMI-IR,* to measure semantic similarity and perform their evaluation on the same 80 question set.

### 4.2 The Results

The *Roget's Thesaurus* system answers 78.75% of the *TOEFL* questions (Table 4). The two next best systems are Hirst St-Onge and *PMI-IR*, which answer 77.91% and 73.75% of the questions respectively. *LSA* is not too far behind, with 64.38%. Bigham e*t al.* (*ibid.*) obtain a score of 97.50% using their combined approach. They further declare the problem of this *TOEFL* set to be "solved". All the other *WordNet*-based measures perform poorly, with accuracy not surpassing 25.0%. According to Landauer and Dumais (*ibid.*), a large sample of applicants to US colleges from non-English speaking countries took the *TOEFL* tests containing these items. Those people averaged 64.5%, considered an adequate score for admission to many US universities.

The *ESL* experiment (Table 5) presents similar results. Once again, the *Roget's* system is best, answering 82% of the questions correctly. The two next best systems, *PMI-IR* and Hirst and St-Onge fall behind, with scores of 74% and 62% respectively. All other *WordNet* measures give very poor results, not answering more than 36% of the questions. The *Roget's* similarity measure is clearly superior to the *WordNet* ones for the *RDWP* questions (Table 6). *Roget's* answers 74.33% of the questions, which is almost equal to a *Good* vocabulary rating according to Reader's Digest (Lewis, 2000-2001), where the next best *WordNet* measure, Hirst and St-Onge, answers only 45.65% correctly. All others do not surpass 25%.

These experiments give a clear advantage to measures that can evaluate the similarity between words of different parts-of-speech. This is the case for *Roget's*, Hirst and St-Onge, *PMI-IR* and *LSA* measures. To be fair to the other *WordNet*-based systems, we decided to repeat the experiments using questions that contain only nouns. The results are presented in Table 7. The *WordNet* measures perform much more uniformly and yield better results, but the *Roget's* system is still best.

## 5 Discussion

We have shown in this paper that the electronic version of the 1987 *Penguin Roget's Thesaurus* is as good as, if not better than, *WordNet* for measuring semantic similarity. The distance measure used, often called edge counting, can be calculated quickly and performs extremely well on a series of standard synonymy tests. Table 8 shows that out of 8 experiments, the *Roget's* is better than *WordNet* every time except on the Rubenstein and Goodenough list of 65 noun pairs.

The *Roget's Thesaurus* similarity measures correlate well with human judges, and perform similarly to the *WordNet*-based measures. *Roget's* shines at answering standard synonym tests. This result was expected, but remains impressive: the semantic distance measure is extremely simple and no context is taken into account, and no word sense disambiguation is performed when answering the questions. Standardized language tests appear quite helpful in evaluating of NLP systems, as they focus on specific linguistic phenomena and offer an inexpensive alternative to human evaluation.

|  | *Penguin Roget* | *WordNet Edges* | *Hirst St.Onge* | *Jiang Conrath* | *Leacock Chodorow* | *Lin* | *Resnik* |
|---:|:---:|:---:|:---:|:---:|:---:|:---:|:---:|
| *Miller Charles* | 1 | 5 | 7 | 6 | 3 | 2 | 4 |
| *Rubenstein Goodenough* | 3 | 5 | 6 | 7 | 1 | 2 | 4 |
| *TOEFL* | 1 | 5 | 2 | 3 | 5 | 4 | 7 |
| *ESL* | 1 | 3 | 2 | 3 | 3 | 3 | 7 |
| *Reader's Digest* | 1 | 3 | 2 | 5 | 3 | 6 | 7 |
| *TOEFL - Nouns* | 1 | 4 | 5 | 2 | 4 | 3 | 6 |
| *ESL - Nouns* | 1 | 3 | 2 | 3 | 3 | 3 | 7 |
| *Reader's Digest - Nouns* | 1 | 4 | 5 | 3 | 2 | 6 | 7 |

**Table 8:** Summary of results – ranking of similarity measures for the experiments

Most of the *WordNet*-based systems perform poorly at the task of answering synonym questions. This is due in part to the fact that the similarity measures can only by calculated between nouns, because they rely on the hierarchical structure that is almost only present for nouns in *WordNet*. The systems also suffer from not being able to deal with many phrases.

The semantic similarity measures can be applied to a variety of tasks. Lexical chains (Morris and Hirst, 1991) are sequences of words in a text that represent the same topic. Links between significant words can be established using similarity measures. Many implementations of lexical chains exist, including one using our electronic *Roget's* (Jarmasz and Szpakowicz, 2003). Algorithms that build lexical chains consider one by one words for inclusion in the chains constructed so far. Important parameters to consider are the lexical resource used, which determines the lexicon and the possible relations between the words, called *thesaural relations* by Morris and Hirst (*ibid.*), the thesaural relations themselves, the transitivity of word relations and the distance — measured in sentences — allowed between words in a chain (Morris and Hirst, *ibid.*). A semantic similarity measure can be used to define the thesaural relations. Our lexical chain building process builds *proto-chains*, a set of words linked via these relations. Our implementation refines the proto-chains to obtain the final lexical chains.

Turney (2001) has used his semantic similarity metric to classify automobile and movie reviews. Bigham *et al.* (2003) use their similarity metric to answer analogy problems. In an analogy problem, the correct pair of words must be chosen amongst four pairs, for example: *cat:meow*: (a) *mouse:scamper*, (b) *bird:peck*, (c) *dog:bark*, (d) *horse:groom*, (e) *lion:scratch*. To correctly answer *dog:bark*, a system must know that a *meow* is the sound that a cat makes and a bark the sound that a dog makes. Both of these applications can be implemented with our version of *Roget's Thesaurus*.

## Acknowledgements

We thank Peter Turney, Tad Stach, Ted Pedersen and Siddharth Patwardhan for their help with this research; Pearson Education for licensing to us the 1987 *Penguin's Roget's Thesaurus of English Words and Phrases*.